# A Novel Visualization System of Using Augmented Reality in Knee Replacement Surgery: Enhanced Bidirectional Maximum CorrentropyAlgorithm

Nitish Maharjan[1,2], Abeer Alsadoon[1,2]*, P.W.C. Prasad[1,2], Salma Abdullah[3], Tarik A. Rashid[4]
[1]School of Computing and Mathematics, Charles Sturt University, Sydney Campus, Australia
[2]Department of Information Technology, Study Group Australia, Sydney Campus, Australia
[3]Department of Computer Engineering, University of Technology, Baghdad, Iraq
[4]Department of Computer Science and Engineering, University of Kurdistan Hewler, Erbil, KRG, Iraq

**Abeer Alsadoon**[1,2]*
* Corresponding author. A/Prof  (Dr) Abeer Alsadoon, [1]School of Computing and Mathematics, Charles Sturt University, Sydney Campus, Australia, [2]Department of Information Technology, Study Group Australia, Sydney Campus, Australia
Email: aalsadoon@studygroup.com, Phone +61 2 9291 9387

## Abstract

***Background and aim:*** Image registration and alignment are the main limitations of augmented reality-based knee replacement surgery. This research aims to decrease the registration error, eliminate outcomes that are trapped in local minima to improve the alignment problems, handle the occlusion and maximize the overlapping parts. ***Methodology:*** markerless image registration method was used for Augmented reality-based knee replacement surgery to guide and visualize the surgical operation. While weight least square algorithm was used to enhance stereo camera-based tracking by filling border occlusion in right to left direction and non-border occlusion from the left to right direction. ***Results:*** This study has improved video precision to 0.57 mm ~ 0.61 mm alignment error. Furthermore, with the use of bidirectional points, i.e. Forwards and backward directional cloud points, the iteration on image registration was decreased. This has led to the improved processing time as well. The processing time of video frames was improved to 7.4 ~11.74 fps. ***Conclusions:*** It seems clear that this proposed system has focused on overcoming the misalignment difficulty caused by the movement of the patient and enhancing the AR visualization during knee replacement surgery. The proposed system was reliable and favorable which helps in eliminating alignment error by ascertaining the optimal rigid transformation between two cloud points and removing the outliers and non-Gaussian noise. The proposed augmented reality system helps in accurate visualization and navigation of anatomy of the knee such as femur, tibia, cartilage, blood vessels, etc.

## 1. Introduction

Knee replacement surgery is transformed from invasive surgery to minimally invasive surgery. It is navigated and visualized by images from Computed Tomography scans, Cone Beam Tomography Scans, Magnetic Resonance Imaging (MRI), X-rays, and image-guided 2D fluoroscopy. Fluoroscopy has been used and considered the standard technique for percutaneous visualization [1]. However, the consistent advancement of 3D imaging and modelling, combined with visualization, has altogether improved the exactness of percutaneous knee surgery [2]. The traditional method based on Computed Tomography (CT) slides does not allow surgeons for the mobility of handling surgical tools in terms of a vision of the internal organs. Similarly, there is a problem in the estimation of the depth of anatomical structures and Cite as : Nitish Maharjan, Abeer Alsadoon, P.W.C. Prasad, Salma Abdullah , Tarik A. Rashid (2020). A Novel Visualization System of Using Augmented Reality in Knee Replacement Surgery: Enhanced Bidirectional Maximum Correntropy Algorithm. The International Journal of Medical Robotics and Computer Assisted Surgery, https://doi.org/10.1002/rcs.2154hand-eye coordination problems while undergoing knee replacement surgery [3]. Surgeons face huge problems while viewing areas of interest clearly and they need to be careful about the risk of cutting other nerves and blood vessels while undergoing the surgery. During intraoperative phases, due to the low quality of images and lack of exact view of intra-operative images, reconstruction of the surgical scene is difficult [3]. This leads to an increase in surgery planning, alignment difficulty, processing time, and depth estimation while attaching artificial knee components [1]. The surgical procedure includes resection, cutting, positioning the implants, resurfacing, and replacing a femoral, patellar or tibial component. The main difficulty in knee surgery is incorrect alignment, which leads to abnormal wear, patellofemoral problems, and mechanical loosening [4]. AR with surgical navigation and visualization are used to solve the above issues for safety and efficiency during surgical procedures [5].

AR is the latest cutting-edge technology that is widely used in the medical field and surgical procedures. They combine the virtual scene with a real-time environment and provides a three-dimensional view of the respective object [5]. AR-based on the medical procedures superimposes the virtual scene onto the patient knee during a medical procedure which furnishes the specialist with the 3D view progressively. AR helps surgeons by providing pragmatic and presumptive information during surgery and guiding the surgeons in complex medical procedures [6]. There are different limitations bing the major issues in this technology such as image registration, depth perception, angular deviation, processing time, and occlusion. So, these limitations should be considered to improve seamless visualization during medical surgery.

In the current context, there are different AR technologies used for surgical guidance such as video-based display, see-through display, and projection-based display [7]. These three technologies are further categorized in marker-based and marker-free registration [7]. Marker-based registration uses bulky trackers and cannot update quickly in real-time. So, to improve surgical flow and minimize the invasiveness during surgery, real-time markerless registration has been used to achieve accuracy and increase the real-time image registration [8]. The use of markerless registration in the navigation system helps to remove the flaw related to the position, angle, distance, vibration of the optical tracker [7]. Different navigation systems have been used to associate the surgical field with their virtual counterparts. Head-mounted display with stereoscopic vision have been used for AR visualization to find the depth capabilities using Microsoft HoloLens [9]. Handling occlusion is also a key aspect of accurate visualization of patient anatomy.  Considering this factor, tracking learning detection (TLD)  was introduced to improve the image registration by using depth mapping based occlusion removable techniques [6]. As image registration is the core part of AR-based surgery, different real- time algorithms such as Stereo Matching algorithm, Iterative closest point, Rotational Matrix and Translation Vector (RMTV), patch-based multi-view stereo (PMVS) algorithm, Coherent Point Drift (CPD), etc. Have been used for accurate and automatic registration.

This paper was designed to enhance the processing time, image alignment, image- registration, video accuracy and eliminate the registration outcomes trapped in local minima. The main aim of this paper is to increase the speed of the alignment process by sampling the data and reducing the number of repetitions for posing refinement. This study purposes a new method to maximize the overlapping parts and to eliminate the registration results which are trapped in local minima by updating the image registration in (Iterative Closest Point)  ICP for better alignment and faster iteration. BiMCC has been used in 3D-ICP registration for ascertaining the optimal rigid transformation between two cloud points. The updated registration method helps to increase the video accuracy where this method is robust to the presence of noise, outliers, and partial difference. Similarly, to remove the occlusion caused by instrument, blood, and by another factor where the occlusion is handled by filling border occlusion in right to left direction and non-border occlusion of the left in the right directions. The remaining sections of the paper are organized as follows: The section Literature Review" describes the range of modules, the section that followed "Methodology and proposed systems" describes the model of the state of the art method, and the details of the proposed model include the block diagram and pseudocode of the proposed formula. The section that followed, "Results and Discussion," discusses the various testing techniques used in this algorithm with different samples. The last section "Conclusion and Future Work" describes the comparison between current and proposed system results and provides the conclusion of the paper.

## 2. Literature Review

A wide range of techniques, algorithms, and modules have been proposed to enhance video accuracy, image registration, and processing time. One of these techniques is a markerless navigation module proposed by Ma, et al. [10]. It has improved the state-of-the-art solution reported previously Chen, et al. [11] They have developed an autonomous surgical method with markerless navigation to help the surgeon in the operating room for enhancing the accuracy and performance in OMS (Oral and Maxillofacial Surgery). The TLD was used to track the bounding box of teeth and the 3D pose was refined by using ICP (Iterative closest algorithm. This system has improved the calculating error and practical error by $0.265 \pm 0.196$ and $-0.069 \pm 0.020$ mm.

To improve accurate localization and intra-operative 3D visualization Rose, et al. [12] have proposed the Augmented Reality system by enhancing the iterative closest algorithm for on-patient medical image visualization. They improvised the state-of-the-art solution of Wang, et al. [13] to visualize the internal anatomic structure of organs to improve both safety and efficiency. Microsoft HoloLens was used to fetch the data from Cross-Platform engine unity and navigational control for simulating and visualizing the surgery. This solution has increased the video accuracy by 2.47 to $\pm 0.46$ mm (in terms of registration error).

De Paolis and De Luca [3] have improved state the art solution of Cleary and Peters [14] by proposing the solution for AR visualization with depth recognition signals that improved the surgeon's presentation in the insignificantly obtrusive medical procedure. They proposed a segmentation and classification algorithm to build the 3D models from medical images and the PivotCalibration algorithm to allow the position and orientation of the tool in a fixed position. In this system, a navigation framework was displayed which supported preoperative and intraoperative stages and provided an X-Ray vision of the patient life structures with depth and distance data. This arrangement gave a very worthy scope of accuracy with the error of roughly 2mm in the movement and position of the surgical tools. Hettig, et al. [15] have improved the state of the art solution of [16] by proposing an Augmented Visualization box (AVB), to diminish the patient distress by performing registration outside the patient's mouth and enhance processing time, Ma, et al. [17]have presented the augmented reality surgical navigation with a precise cone-beam computed tomography (CBCT) patient registration method. They improved state the art solution of Hassfeld and Mühling [18] which increased both dental implant accuracy and registration accuracy of the system by enhancing registration between pre-operative data and intra-operative data using a singular value decomposition algorithm and registration devices. It has provided an accuracy of 1.2mm (target registration error) and 4.03° (angular error). However, registration accuracy and implant accuracies have been improved.

Zhang, et al. [19] have improved the state-of-the-art solution of Dimick and Ryan [20] and Souzaki, et al. [21] by enhancing markerless deformable image registration procedures with an ICP algorithm to minimize the target registration error. A coarse to fine deformable image registration was done between preoperative data and intraoperative data using iterative closest point algorithm followed by CPD algorithm to enhance video accuracy and registration between two cloud points. This work has provided the registration error by $1.28 \pm 0.68$ mm (root mean square error) and improved the automatic segmentation by 94.9, Meng, et al. [22] improved the state of the art solution of Taylor, et al. [23] by introducing a multiview stereo vision algorithm to capture different angles around patient heads by the movement of a robot end-effector with an optical camera. It has provided a target registration error from 1.34 mm to 2.33 mm without two outliers.

Burström, et al. [1] have improved state of the art solution of Kim, et al. [24] by enhancing the video accuracy and depth perception with the help of automatic segmentation and pedicle identification algorithm. They proposed ARSN (Augmented Reality Navigation) This system has automatic spine calibration and pedicle identification algorithm to plan, guide, and insert the pedicle instrument. This solution increases the navigation and technical accuracy by 1.7 ±1.0 mm at the bone entry point and 2.0±1.3 mm at the device tip, clinical accuracy by 97.4% to 100% according to screw size. Similarly, Jiang, et al. [25] have improved state of the art solution of Fortin, et al. [26] by introducing the marker-less point-cloud image patient registration method. This method merges the virtual images in the actual environment for dental implant surgery using a 3D image see-through overlay. This system offers good video accuracy with minimal registration error and angular deviation. This solution provides a registration error of 0.54 mm with the < 1.5 mm mean linear deviation and < 5.5-degree angular deviation.

Gerardo, et al. [27] Have improved state of art solutions of Cassetta, et al. [28] by introducing the ImplaNav navigation system for dynamic guidance with calibration and refinement of HoloLensto take control and show multi-dimensional images or virtual images and visualize 2D/3D information of bone structure by intra-oral navigation., it has produced a video accuracy as 0.46mm by the entry point and 0.48mm at an apical point at the apical point. Similarly, Talaat, et al. [9] Proposed research to improve the performance of traditional orthodontic treatment by using the Microsoft HoloLens with OMSA algorithm for 3D dental models They have improved virtual assessment limitations of Hirogaki, et al. [29] for orthodontic and oral surgery. The superimposition was achieved by OMSA algorithm and concordance correlation coefficients were calculated by the lins algorithm to calculate the measurement error. It has improved the real 3D assessment of orthodontic tooth movement and achieve an accuracy of 0.5 mm.

Murugesan, et al. [8] have improved state of the art solution of Wang, et al. [30] by improving ICP with the RMTV algorithm to reduce the geometric error. They have used two stereo cameras to improve depth perception. Among the six stages of ICP, the error metric was focused on reducing the overlapping error. With the help of the proposed algorithm, video accuracy was improved by 0.3-0.4 mm overlay error and processing time of 7-10 frames per second. To eliminate occlusion and a noises in the pictures. Basnet, et al. [6] Have enhanced the TLD algorithm by introducing a noise removable technique for better tracking of the surgical area and reducing the processing time of real-time videos. Modified kernel non-local means (MKNLM) have been used to remove noise from real-time videos which can negatively impact image registration and deteriorate the image edges. The solution had a very acceptable overlay accuracy of 0.23-0.35 mm and a processing time of 6-11 frames per second. Pokhrel, et al. [5] Have proposed a volume subtraction method to analyze the cutting areas to improve the alignment accuracy and processing time. This system used the differential information for finding the target shape and current shape to detect the alignment and remaining area need to cut. The remaining area to be cut was ascertained using Markov random field surface reconstruction algorithm. This solution increases the accuracy by 0.40-0.55 mm and reduces the image processing time from 12-13 frames per second to 9-10 frames per second. the state of the art solution of Wang, et al. [30] was improved by recent work [31] to enhance the image registration between two cloud points and depth perception. The movement of the patient during image registration was also solved this system furnishes a satisfactory scope of precision with a minimum processing time, which gives the surgeon a smooth careful visualization. It provides a target registration error less than 0.50 mm (overly error) with 0.5s processing time.

3. **Methodology and proposed systems**

**3.1 State of the Art**

Wang, et al. [31] have proposed an ICP with a stereo matching algorithm and Gaussian filter to enhance the alignment, position and remove the geometric error. This paper explained how the list of images was created by a hierarchical model which is transferred to Stereo Based Tracking to which tracks the surgical area with the help of a bounding box. Then, overlaid 2D image inside the bounding box is re-examined and registration is done by ICP. Further, refinement and elimination of the geometries are done by Stereo

Matching Algorithm. Gaussian Filter is used to removing the noise caused by camera calibration and other registration procedures. It provides a target registration error of less than 0.50 mm with 0.5s processing time.

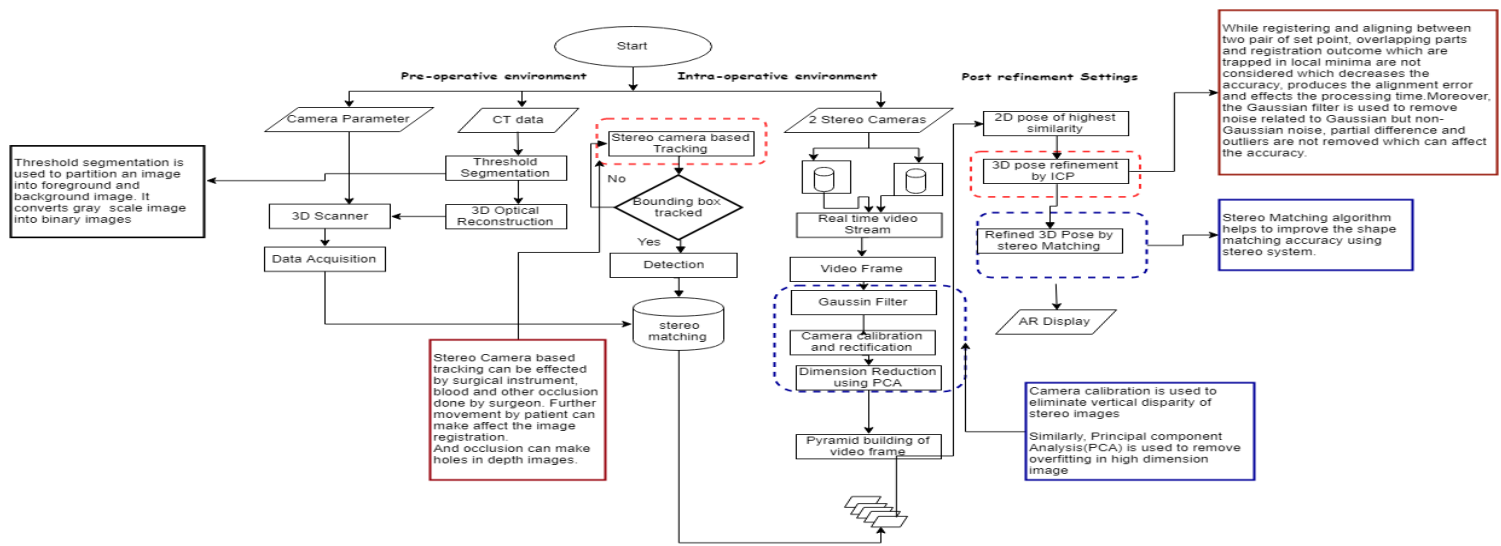

Fig. 1: Block Diagram of the State of the Art the System, [31]
[The blue borders show the good features of this state of the art solution, and the red border refers to the limitation of it]

This model consists of three components as shown in (Fig 1: State of the Art) which are a pre-operative environment, intra-operative environment, and post-refinement settings.

*Pre-operative Environment*: In the pre-operative environment, (CT) scan of a patient is acquired before undertaking the surgery. The CT data are segmented for obtaining precise surgical areas. Threshold segmentation is done to partition an image into the foreground and background image. The 3D scanner is used to obtain the 3D model data of teeth to find virtual information.

*Intra-operative Environment:* In Intra-operative processing, two stereo cameras are used to capture a high -quality large images. This stereo camera records two different angles views of the same scene. The main advantage of using these images is to find depth information. The distance between one image to another is merging into another image and attempts to map every pixel in one image to a location in another image and helps to find the disparity between two images. Then, CT derived model is matched with a stereo camera. Then, a 3D scanner is used to remove the shape inconsistency while matching with pre-operative data. Furthermore, the 3D model generated from the 3D scanner is used for the 3D reconstruction of teeth. Then, the camera is calibrated to reduce the vertical disparity of images. Then, the 3D model generated from the 3D scanner is matched with pre-operative CT data. Then, principal component analysis is used to remove overfitting in high dimension images. So, reconstructed exposed teeth based on CT data are matched with stereo cameras. With the use of the ICP algorithm, the registration and tracking are done with a stereo camera in this phase. Then, real-time surgical video is tracked by a stereo camera for tracking the position of the surgical area with the help of a bounding box. Bounding helps to lessen the search area and remove the false results. Then the result is matched with a list of segmented images from offline or online matching and the 2D image is overlaid onto the video frame. The comparison of two rendered aspects of the camera is used to extract the 2D projected shapes and an average absolute value of the dot product of the direction vectors is calculated on the overlapped components. The aspect with the highest similarity score is tracked and selected. Further, refinement is done by the stereo matching algorithm to enhance the alignment and position. Then, a Gaussian filter is used to eliminate the noise, roughness, and waviness from the primary surface.

The above state of the art helps to match pre-operative data from exposed teeth model and stereo image. But it does not handle the occlusion handling. Stereo-based tracking cannot handle occlusion from instruments and blood. This research has not explained this feature. Occlusion handling should be added to this system because while tracking the real surgical l image from a stereo camera, the surgical site can be blocked from a stereo camera. Similarly, the images cannot be visualized, and the surgical site cannot be tracked by a stereo camera which leads to registration error and makes holes in in-depth images. The main limitation of this research is in registering and aligning two different cloud points. So, while registering and aligning between two pairs of set points, overlapping parts and registration outcomes trapped in local minima are not considered which decreases the accuracy, produces the alignment error, and affects the processing time.

*Pose Refinement Settings:* In this process, the Iterative closest point refines the pose and overlaid it into the surgical area. The Stereo Matching algorithm refined the images and the Gaussian filter helps to remove the geometric error. A stereo camera is used to project the surgical scene onto a translucent mirror which is placed above the surgical area and viewed by surgeons while undertaking the surgery Wang, et al. [31].

The rigid transformation which satisfies the best correspondence between two given cloud points to minimize the error using ICP is defined in equation (1) [31].

$$E(R, t) = \sum_{i=1}^{N} [(R_{pi} + \vec{T}) - \overrightarrow{m_{c(i)}}]^2 \qquad (1)$$

Where,
$P \triangleq \{\vec{p}_i\}_{i=1}^{N_p} (I \ (N_p \in N)$ is the shape point set
and $M \triangleq \{\vec{m}_j\}_{j=1}^{N_m} (N_m \in N)$ is the model point set,
N is the number of paired points,
$R \in R^{n*n}$ is the rotation matrix,
$\vec{T} \in R^n$ is the translation vector,

The above algorithm is being used to search the optimal mapping which is composed of translation and rotation to reduce the distance between two points in the least-square sense.

The model's pose is determined from the viewpoint and the maximization problem is solved by following equation (2) [31]: -

$$M(I) = \frac{max}{t}\frac{1}{2}(s\ (Proj(obj, P_l T), I_l) + s\ (Proj(obj, P_r T), I_r)) \qquad (2)$$

Where,
    $Proj(obj, P_l T), I_l)$, i = lr is the 2D projected shape of the 3D model obj using the projection matrix
    $P_l = K(I, 0)\ and\ P_r = K(I, b)$ is the projection matrix of left and right cameras
    $P_i T$, i= lr is the projection matrix
    T= (R, t; 0, 1) is the pose of the obj with respect to the left camera;
    s (·, ·) is the metric measuring the similarity between the projected 2D shape and the image;
    Il and Ir represent the left and right image of the camera.
    I is the image

Similarity, metric between 2D projected shape with N points and an image I are defined as follows in equation (3):

$$S = \frac{1}{N}\sum_{i=1}^{N} \frac{|d_i^T \nabla I(x_i,y_i)|}{||\nabla I(x_i,y_i)|| \, ||d_i||} \qquad (3)$$

Where,

$\nabla I(x_i, y_i)$ represents the image gradient at $(x_i, y_i)$.
$(x_i, y_i)$ : projected 2D shape of 3D model
$d_i^T$ : direction vector
, : Denoted dot product
$N$ : number of feature points
s is the metric measuring the similarity between the projected 2D shape and the image;

Refinement of the image was obtained done by aligning the 3D model with stereo images as follows in equation (4) [31].

$$R(I) = \frac{min}{Rt} \frac{1}{N_l+N_r} \left( \sum_{x_i^l \in r^l} dist(K(RX_i^l + t), X_i^l)^2 + \sum_{x_i^r \in r^l} dist(K(RX_i^r + t + b), X_i^r)^2 \right) \qquad (4)$$

Where,

dist (x, y) represents the Euclidean distance between the inhomogeneous coordinates of x and y;
$N_l$ and $N_r$ are the point number of $l^l$ and $I^r$ , respectively.
R: represent rotation matrix
T: represents translation vector
$x_i^l \in r^l$ is selected to minimize $RX_i^l$ along with the search direction K
$X_i^l$ is the scalar parameter
N: number of feature points
I is the image

| **Algorithm 1: Enhanced ICP with Stereo Matching Algorithm** |
|---|
| Input: CT data, and optically calculated data |
| Output: Augmented image |
| BEGIN |
| Step 1: During the preoperative stage, CT data of the patient are acquired, and the 3D scanner is used to generate a 3D model of teeth, then threshold segmentation is done to partition an image into a foreground image.<br>Two Stereo Camera are used to capture the right and left image and used to capture video stream<br>and Camera calibration and rectification are done to eliminate vertical disparity of images.<br>During surgery, a stereo camera is used to track the surgical site using the bounding box. Then ICP is used for registration between CT data and intraoperative data using the following equation:<br>$\quad$ E (R, t) $=\sum_{i=1}^{N}[(R_{pi} + \vec{T}) - \overrightarrow{m_{c(i)}}]^2$ $\qquad$ (1)<br>Step 2: The surgical video from the stereo camera is matched with preoperative data and refine by a stereo matching algorithm.<br>Step 3: In the stereo matching algorithm, matching cost computation is the first step where preliminary data of stereo matching differences of every pixel is calculated. This stage must be good and robust to avoid noise. A sum of gradient matching is introduced using fixed window size.<br>$\quad$ SG(x,y,d)$=\frac{1}{w^2}\sum_{(x,y,d)\in w_{sg}}|m_l(x,y) - m_r(x,y,d)|$ $\qquad$ (2)<br>Where (x,y) is the pixel of interest coordinates, d represents the disparity value, $w_{sg}$ represents the window size, $m_l$ and $m_r$ are magnitude values.<br>Step 4: Cost aggregation is done which is the second step of stereo matching. This stage helps to eliminate the noise from matching cost computation and make the result more consistent. The spatial distance between the two points is determined.<br>Step 5: Disparity Selection and Optimization is done to minimize the data selection on a specific location and represent with disparity value.<br>$\quad$ d = $\arg min_{d\in D} CA(p,d)$ $\qquad$ (3)<br>where D is the set of disparity value of an image<br>Step 5: The last stage of the stereo matching algorithm is a post-processing and refinement stage where invalid disparity value is managed, and the remaining noise is removed. It is done by left-right consistency checking process.<br>Step 4: Gaussian filter is used to remove the noise<br>Step 5: Augmented image is projected<br>END |

**3.2 Proposed System**

Different methods of Augmented reality-based surgery are reviewed, different, so different features were analyzed and limitation of each method was proposed by given authors. From analyzing those methods, it was found that there were different issues related to AR-based surgery. Such as image registration, cutting errors, video accuracy, processing time, occlusion, etc. Therefore, it was selected the [1] state of the art, that based on [31] and BiMCC was used in ICP to improve the registration process. The maximum Correntropy was introduced in ICP for eliminating the outliers and non-Gaussian noise. This algorithm was used instead of Euclidean distance to remove outliers and noise. Similarly, this algorithm helps to maximize the overlapping parts and eliminate the registration outcomes trapped in local minima (darkest pixel). This algorithm is simple, neat, and effective and works better than feature-based methods and normal-based methods in terms of convergence and robustness. This solution makes this algorithm more robust and performs better in noisy conditions. Due to its simplicity and low computational cost, it seems like one of the most popular techniques. BiMCC was used to increase the speed of the alignment process of the data sampling. The goal of this solution is to search the correspondence and ascertain the optimal rigid transformation between two cloud points which helps to increase the accuracy and processing time. Also, eliminating the noises in the registration phase helps to enhance the speed of the algorithm compared to traditional ICP [32].

Along with the above solution, the use of Weighted Least Square (WLS) in stereo vision tracking helps to remove the occlusion by filling border occlusion in right to left direction and non-border occlusion of the left to the right direction [33]. It is hard to get disparity information from occluded pictures. So, WLS first selects the occluded point, defines the neighborhood of occluded point, and interpolated by predicting the unknown values for those occluded points, as shown in Fig.2.

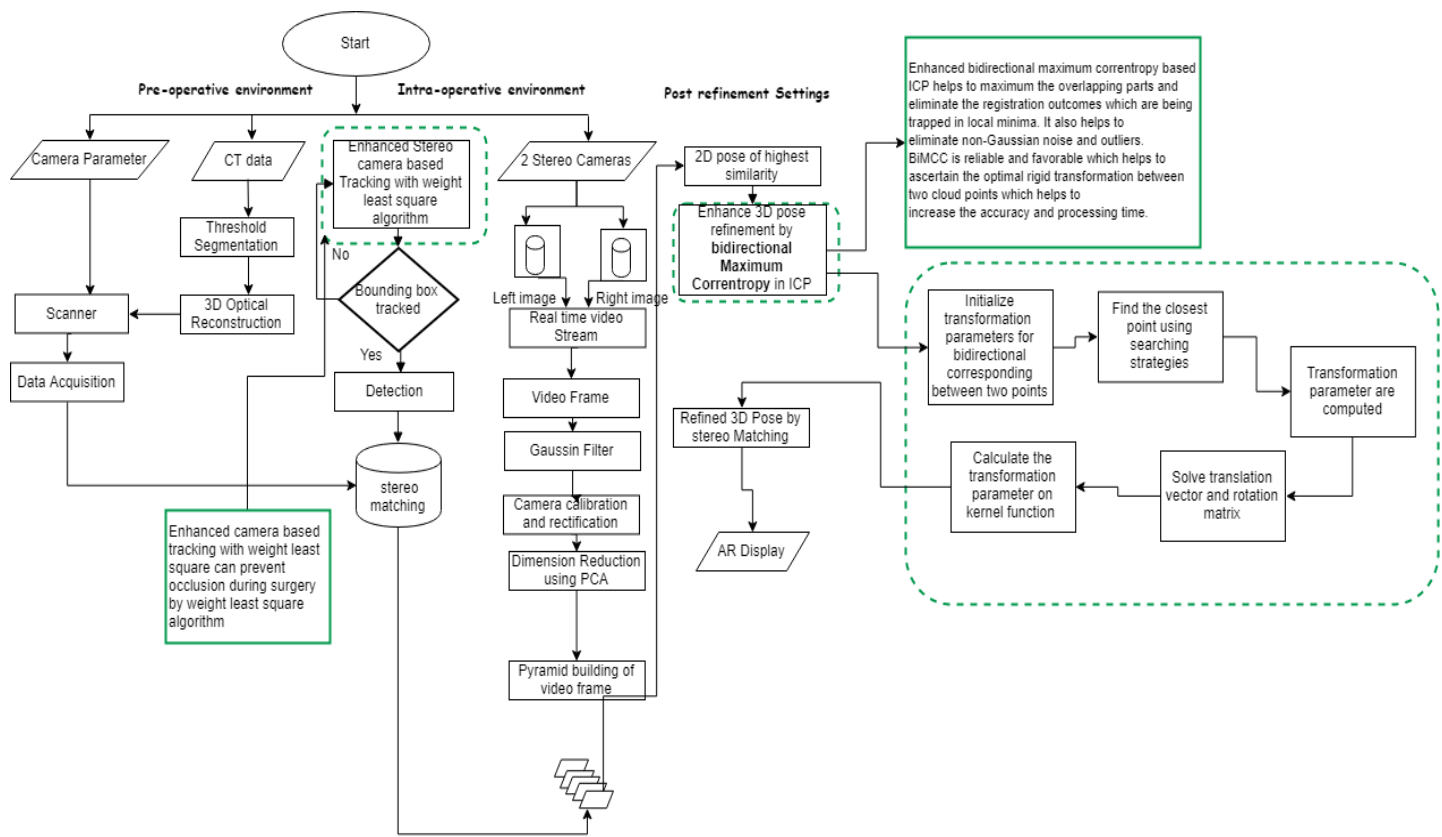

Fig. 2. Block diagram of the proposed AR system for knee replacement surgeries using Bidirectional Maximum Correntropy Algorithm
[The green color refers to the added new parts by us]

***Pre-operative Environment:*** In this environment, computed tomography (CT) scan of a patient is acquired before undertaking the surgery. The CT data were segmented to obtain precise surgical areas. Threshold segmentation is done to partition an image into a foreground and background images. The image reconstruction is carried out by finding the region of interest and surgical area of the knee is planned based on the CT image.

***Intra-operative Environment:*** In the environment, two stereo cameras were used to capture the high-quality large images. Two stereo cameras produced the left and right images of the surgical site. This stereo camera records two different angle views of the same scene. The main advantage of using these images is to find the depth information. Firstly, it finds the distance between one image with another image and attempts to map every pixel in one image to location of another image. This process helps to find the disparity between the two images. Then, CT derived model is matched with the model with stereo camera to remove the shape inconsistency while matching with pre-operative data. Then, the camera is calibrated to reduce vertical disparity of images. Two stereo cameras were used to capture the surgical procedures in real-time and this is sent to a stereo-based tracking algorithm to track the location of the surgical area with the help of a bounding box. In stereo-based tracking, occlusion is not handled and removed. So, we have applied the weight least square algorithm which will handle the occlusion by filling border occlusion from right to left direction and non-border occlusion from left to right direction. This algorithm will first select the occluded point, define the neighborhood of the occluded point and interpolate by providing or predicting the unknown values for those occluded points. Then the result was matched with a list of segmented images from offline or online matching and inside the bounding box. 2D image is overlaid onto the video frame. The similarity of the two rendered views is generated from the camera by extracting the 2D projected shapes and then calculating the average absolute value of the dot product of the direction vectors on the overlapped pixels. The aspect with highest similarity score is tracked and selected. Further, refinement is done by stereo matching algorithm to enhance the alignment and position. Then, a Gaussian filter is used to remove the noise, roughness, and waviness from the primary surface. Initial alignment is done by matching data generated from the pre-operative and stereo camera. The further registration is done in the post-refinement phase.

*Pose Refinement Settings:* In this environment, (BiMCC) in 3D-ICP registration is used to improve the ICP and registration process. It helps to enhance the registration process and eliminates alignment error by removing the outliers and non-Gaussian noise. BiMCC is reliable and favorable which helps to correspondence and ascertain the optimal rigid transformation between two cloud points which helps to improve the accuracy and processing time. Similarly, Parameter estimation is solved to speed up computation and convergence by using two set points from the current correspondence of x and y cloud points. Then, the Stereo Matching algorithm [31] refines the images and the Gaussian filter helps to remove the geometric error. A stereo camera is used to project the surgical scene onto a translucent mirror which is placed above the surgical area and viewed by surgeons while undertaking the surgery [31].

### 3.1 Proposed Equation

The rigid transformation which satisfies the best correspondence between two cloud points to minimize the error by using ICP is formulated in equation (1) [31].

$$E(K) = \sum_{i=1}^{N} [(R\vec{x_i} + \vec{T}) - \vec{y}_{c(i)}]^2 \tag{1}$$

However, above ICP algorithm has simplicity and clarity but it needs modification and improvement by using all six variants to prevent the outcomes trap in local minima.

Where,

$x \triangleq \{\vec{x_i}\}_{i=1}^{N_x}$ ($N_x \in N$) is the shape point set
and $y \triangleq \{\vec{y_j}\}_{j=1}^{N_m}$ ($N_y \in N$) is the model point set,
N is the number of paired points,
$R \in R^{n*n}$ is the rotation matrix,
$\vec{T} \in R^n$ is the translation vector,
$\vec{x_i}$ and $\vec{y}_{c(i)}$ are the two corresponding cloud points.

The above algorithm is composed of translation and rotation vectors which help to reduce the distance between two points in the least-square sense. The registration process is the rigid transformation between point $\vec{x_i}$ by rotation matrix and translation vector $\vec{T}$ to $\vec{y}_{c(i)}$.

The error between the two above matched points is given in equation (2):

$$e_{xi} = (R_{xi} + \vec{T}) - \overrightarrow{y_{c(i)}} \tag{2}$$

Without a good initial position of x, the position of y will be wrong at the beginning. So, it is needed to fix the rotation matrix and then other translation points. By computing equation 1 and 2, modified Rigid transformation in ICP is formulated in equation (3): -

$$E(K) = \sum_{i=1}^{N}[exi]^2 \tag{3}$$

To speed up the computation and convergence, we have used two set points $A = \{\vec{a_i}\}_{i=1}^{N}$ and $B = \{\vec{b_i}\}_{i=1}^{N}$ from current correspondence of x and y cloud points were used. These points are used to solve parameter estimation[32].

These two points can be defined in equation (4) : -

$$\vec{a_i} = \begin{cases} \vec{x_i}, & 1 \leq i \leq N_x, \\ x_{d_k(i)}, & N_x + 1 \leq i \leq N. \end{cases}$$

$$\vec{b_i} = \begin{cases} y_{c_k(i)}, & 1 \leq i \leq N_x, \\ \vec{y_i}, & N_x + 1 \leq i \leq N. \end{cases} \tag{4}$$

Where,
$\vec{x_i}$ and $\vec{y}_{c(i)}$ are the two corresponding cloud point
$c(i)$ is the corresponding index
$d(k)$: is the point sets
N is the number of paired points
$A = \{\vec{a_i}\}_{i=1}^{N}$ and $B = \{\vec{b_i}\}_{i=1}^{N}$ are two sets points

Using two sets of points from the current correspondence of x and y cloud points, computation and convergence time are increased. It was behaving derived modified error between two matched points in equations (5) and (6).

$$Me_{xi} = (R\vec{a_i} + \vec{T}) - \vec{b_i} \tag{5}$$

$$Me_{yi} = (R\vec{b_{d(j)}} + \vec{T}) - \vec{b_j} \tag{6}$$

Where,
$A = \{\vec{a_i}\}_{i=1}^{N}$ and $B = \{\vec{b_i}\}_{i=1}^{N}$ are two sets points
d(j): distance between two points
R is the rotation matrix,
$\vec{T}$ is the translation vector,
$\vec{x_i}$ and $\vec{y}_{c(i)}$ corresponding cloud points are transformed into $\vec{a_i}$ and $\vec{b_{d(j)}}$ to speed up the computation because it consists of good searching strategies.

The Bidirectional maximum correntropy used forward and backward direction during image registration, which helps to decrease the iteration during registration and increase video accuracy. The bidirectional maximum correntropy in ICP is formulated in equation (7) [32]: -

$$\hat{v}_\sigma(\{X,\tilde{X}\},\{Y,\tilde{Y}\},R,\vec{t}) = \frac{1}{N}(\sum_{i=1}^{N_x} \exp(\frac{-||R\vec{x_i}+\vec{t}-\vec{y}_{c(i)}||^2}{2\sigma^2}) + \sum_{j=1}^{N_y} \exp(\frac{-||R\vec{y_{d(j)}}+\vec{t}-\vec{y_j}||^2}{2\sigma^2})) \tag{7}$$

Where,
The above equation is the sum of forwarding distance and backward distance of two point sets,
X and Y are two cloud points,
$\tilde{X}$ and $\tilde{Y}$ are two corresponding points,
$\{X,\tilde{X}\},\{Y,\tilde{Y}\}$: two big point sets,

$\vec{x_i}$ in the opposite cloud point is $\vec{y}_{c(i)}$,
R is the rotation matrix,
$\vec{t}$ is the translation vector,
σ denotes kernel bandwidth,
$N=N_x+N_y$: Same point number,
c(i) : one-to-one corresponding index,
d(j): the distance between two points

By computing equation 3,5,6, and 7, modified bidirectional maximum correntropy in ICP as shown in equation 8,9,10: -

$$M(\hat{v}_\sigma) = \frac{1}{N} \left( \sum_{i=1}^{N_x} \exp\left(\frac{-||R\vec{a_i} + \vec{T}) - \vec{b_i}||^2}{2\sigma^2}\right) + \sum_{j=1}^{N_y} \exp\left(\frac{-||R\vec{a_{d(j)}} + \vec{T}) - \vec{b_j}||^2}{2\sigma^2}\right) \right) \tag{8}$$

The Modified bidirectional maximum correntropy in ICP is formulated as [32]. The following equation is computed by updating the error function from equation 5,6 and 8: -

$$M'(\hat{v}_\sigma) = \frac{1}{N} \left( \sum_{i=1}^{N_x} \exp\left(\frac{-||Mexi||^2}{2\sigma^2}\right) + \sum_{j=1}^{N_y} \exp\left(\frac{-||Meyi||^2}{2\sigma^2}\right) \right) \tag{9}$$

Where,
The above equation is the sum of forwarding distance and backward distance of two points sets,
X and Y are two cloud points,
$\tilde{X}$ and $\tilde{Y}$ are two corresponding points,
$\{X, \tilde{X}\}, \{Y, \tilde{Y}\}$: two big point sets,
$\vec{x_i}$ in the opposite cloud point is $\vec{y}_{c(i)}$,
R is the rotation matrix,
$\vec{t}$ is the translation vector,
σ denotes kernel bandwidth,
$N=N_x+N_y$: Same point number,
c(i) : one-to-one corresponding index,
d(j): the distance between two points

The registration outcome trapped inside local minima and convergence is improved from the following equation (10). The equation is derived from equation 1 and equation 9: -

$$ME\ (K) = \sum_{i=1}^{N} \exp[Mexi + Meyi]^2 / 2\sigma^2 \tag{10}$$

Where,
$Me_{xi} = (R\vec{a_i} + \vec{T}) - \vec{b_i}$,
$Me_{yi} = (R\vec{b_{d(j)}} + \vec{T}) - \vec{b_j}$
R is the rotation matrix,
$\vec{T}$ is the translation vector,
$\vec{x_i}$ and $\vec{y}_{c(i)}$ corresponding cloud points are transformed into $\vec{a_i}$ and $\vec{b_{d(j)}}$ to speed up the computation because it consists of good searching strategies.
$\vec{x_i}$ in the opposite cloud point is $\vec{y}_{c(i)}$,
R is the rotation matrix,
$\vec{t}$ is the translation vector,

σdenotes kernel bandwidth, which helps for scalability
c(i): one-to-one corresponding index,
d(j): distance between two points

## 4. Results and Discussion

The implementation of our proposed system was done by a multi-paradigm numerical processing environment called MATLAB R2019b. Using this proprietary language, 17 knee surgery videos and 17 simple CT scan images of different patients were taken to analyze and design the process with the help of array mathematics. The knee replacement video was taken as sample where knee components can be easily distinguished such as tibia, femur, patella, and cartilage. Sample video and CT scan data play an important role while undertaking this experiment. The video was extracted from the internet which varies from 10-20 minutes with frame rate 24fps and 30fps. However, videos are long and consist of unwanted frames and noisy data where they have selected the part of videos that display femoral component, patellar component, and a tibial component. The picture edges have been extricated utilizing MATLAB, and every image was considered. The separated picture outlines of every video are shifted depending on the video length. The sample video starts with cutting with shin bone, ligament and meniscus and replacing with artificial joints. The image quality depends on algorithms, pre-operative data, and stereo cameras. We have created all of the figures.  We used YouTube open-source videos from which the screenshots of the breast images were taken. The table in which those images appear was also created by us.  As noted within the manuscript, our study did not involve human participation; therefore, as per the Australian Code for the Responsible Conduct of Research and the National Statement on Ethical Conduct in Human Research, ethics committee approval for this project was not sought. For this reason, no such details were mentioned in the Methods. We have now mentioned this point in the Methods section as recommended by the reviewer.

By using the proposed system shown in Fig 3, it was found great results between state of art in Fig 3 (b) and proposed system in Fig 3 (c). Alignment problems and registration errors have been improved than the previous state-of-the-art system. Processing time has also been improved as shown in  Fig 5. The outliers and noises presented in a state of art have been eliminated. The overlapping problems between two cloud points and depth perception have also been removed.

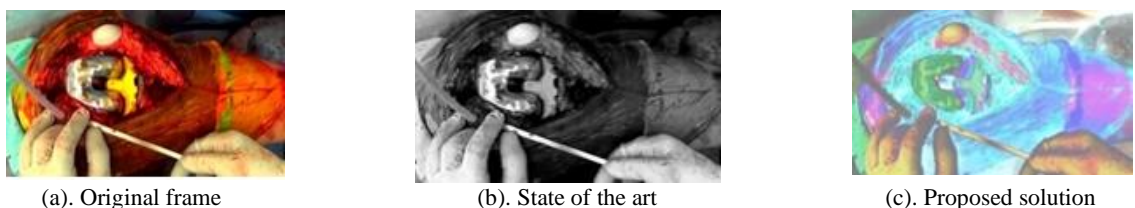

(a). Original frame                 (b). State of the art                 (c). Proposed solution
Fig 3: Comparison of result in a state of the art and proposed solution

Various functions have been used to measure code execution time by calculating two instances of time. One of the major functions used in this system is the performance measurement function which is used to measure the accuracy of the system. Performance time is measured by using time it () function which helps to find the number of times to run the code in the system. The time it () functions help to call the function numerous times and returns the median of the measurements. Processing time is calculated by number of frames accumulated by live video from a stereo camera per second during matching of real-time video. Video accuracy and processing are related and can be ascertained using the above functions. Aspect graphs and disparity maps are used to make registration and matching of specified objects in real-time videos from different angular perspectives.

Cite as : Nitish Maharjan, Abeer Alsadoon, P.W.C. Prasad, Salma Abdullah , Tarik A. Rashid (2020). A Novel Visualization System of Using Augmented Reality in Knee Replacement Surgery: Enhanced Bidirectional Maximum Correntropy Algorithm. The International Journal of Medical Robotics and Computer Assisted Surgery, https://doi.org/10.1002/rcs.2154After successfully extracting pre-operative CT data, the surgeon uses different surgical materials and techniques in the intra-operative environment for effective procedures to remove infection and to correct leg deformity. In the intra-operative stage, firstly surgeon finds out the damaged's cartilage from the CT data, then the incision is done with the help of stereo-based tracking. Two stereo cameras are used for binocular vision. It is similar to human eyes which help to find the depth perception while undertaking the surgery. Two stereo cameras help to record the surgery video and overlay images over the patient organ.

After partitioning an image into foreground and background images, threshold segmentation was used in a pre-operative environment. Then grayscale image was converted into binary images. The outcome was contrasted with the fragments of the images from pre-operative CT pictures for the visualization of knee replacement procedure from bone penetrating to arrangement of prosthetic. Implementation of the proposed system from finding the damaged cartilage, position of metal implants, resurfacing the patella and insertion are defined, as shown Fig 4 below:

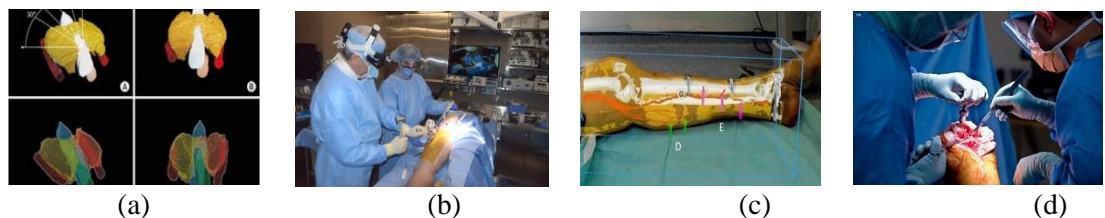

(a)      (b)      (c)      (d)

Fig 4: Implementation of Proposed System

(a) Image registration (b) Stereo matching algorithm with stereo based tracking (c) Image alignment with correntropy algorithm (d) Drilling bone using image overlay

Comparison between state-of-the-art and proposed solutions was derived with the assistance of diagrams and information reports. With the data of 17 patients, proposed solutions were experienced in different stages such as bone preparing , finding the damaged knee components, positioning the implant, drilling of knee bone, and resurfacing the patella. The surgical experiment was affected by different impulse noise, outlier, patient movement and occlusion. The result was calculated in terms of image registration, image alignment, and processing time. Video accuracy of proposed solutions was calculated in terms of alignment error and overlay error. It was carried extensive test for 20 tests, 5 tests in every situation such as opening knee bone after locating damaged cartilage, drilling knee bone, positioning of implant, resurfacing of patella and positioning of metal implants. Those 5 tests had 4 different scenarios which may affect the video accuracy if there is movement of the patient's body or by movement of instruments. The video accuracy has been determined by taking the normal after effect of each test cases. At that point the conclusive outcome was determined by taking average for all experiments in five situations.

These outcomes were analyzed during various phases of AR-based knee replacement surgery. The proposed system has an enhanced image registration, improved the misalignment difficulties faced by patient and eliminated the registration outcomes trapped in local minima. The proposed arrangement has improved the video accuracy by ascertaining the optimal rigid transformation between two cloud points and removing the outliers and non-Gaussian noise. The proposed augmented reality system has helped in accurate visualization and navigation of the anatomy of the knee such as femur, tibia, cartilage, blood vessels etc. Furthermore, this system has improved the processing time by decreasing the number of iterations during image registration using bidirectional point i.e. Forwards and backward directions in ICP. The proposed system has been implemented in intra-operative stages after partitioning an image into foreground and background image by extracting pre-operative data.

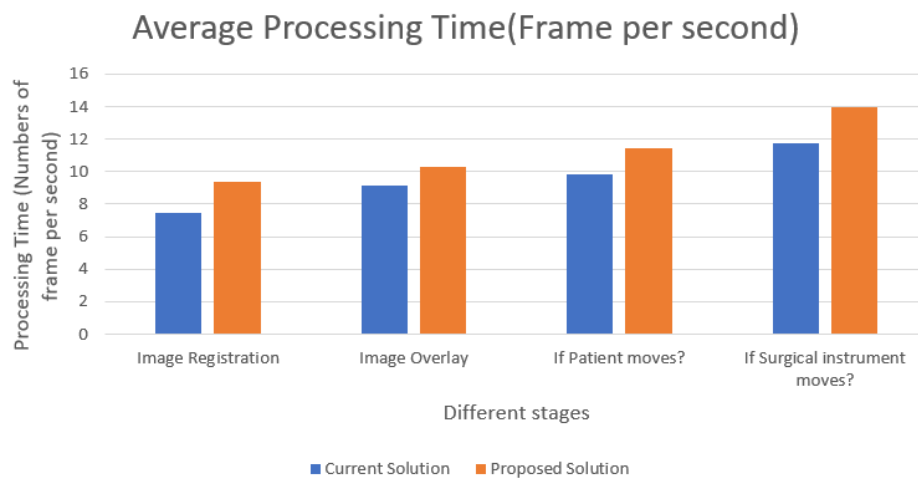

Fig 5: Average processing time results for proposed and state of the art solutions by registration and alignment error

Figure (5) shows the variance between state of the art and proposed solutions in terms of processing time. The first couple bar represents the average processing time by alignment error during image registration, the second couple bar represents the average processing time by alignment error during image overlay, the third couple bar represents the average processing time of alignment error during movement of the patient and last couple bar represents the average processing time by alignment error during instrument movement.

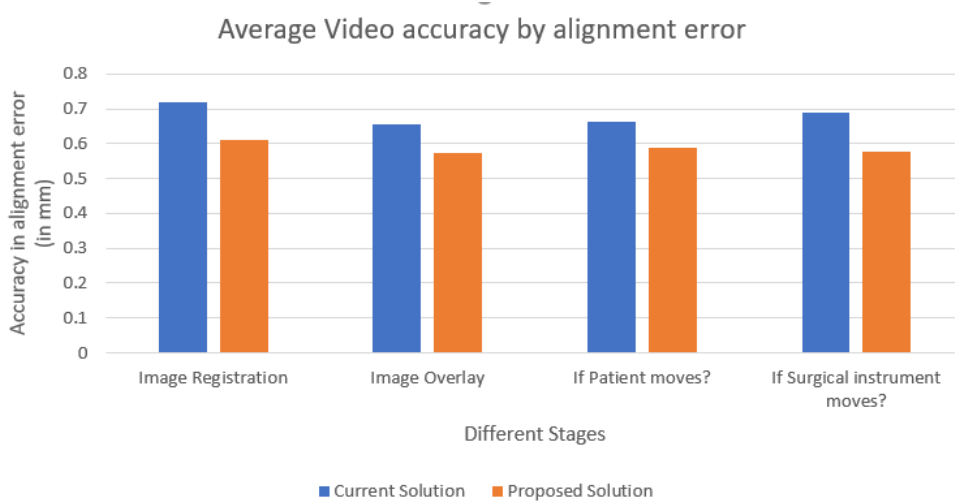

Fig 6: Average video accuracy result of  proposed and state of the art solutions by registration and alignment error

Figure 6 displays vary between state of the art and proposed solutions in terms of alignment error. The first couple bar represents the average video accuracy by alignment error during image registration, the second couple bar represents the average video accuracy by alignment error during image overlay, the third couple bar represents the average video accuracy by alignment error during movement of the patient and the last couple bar represents the average video accuracy by alignment error during instrument movement.

**Table 1: Accuracy and Processing Time Results for Knee replacement (Sample 1: Adult man Age 35 to 45)**

| Sample 1 Stages | Sample details Adult man Age 35 to 45 | Original video | State of the Art | | | Proposed solution | | |
|---|---|---|---|---|---|---|---|---|
| | | | Processed sample | Accuracy by alignment error | Processing time (Frames per second) | Processed sample | Accuracy by overlay error | Processing time (Frames per second) |
| 1.1 | Opening Knee Bone (Man age – 35) | Image registration 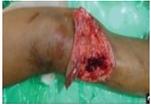 | 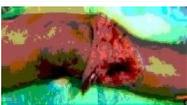 | 0.65 mm | 7 fps | 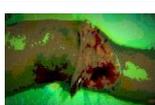 | 0.52 mm | 15 fps |
| | | Image overlay 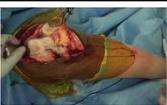 | 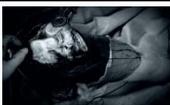 | 0.46 mm | 9 fps | 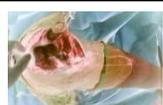 | 0.40 mm | 10 fps |
| | | If patient moves? 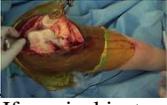 | 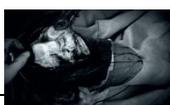 | 0.50 mm | 10 fps | 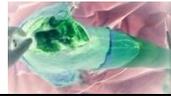 | 0.40 mm | 12 fps |
| | | If surgical instrument moves? 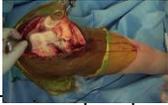 | 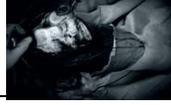 | 0.61 mm | 10 fps | 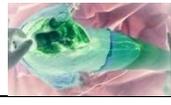 | 0.55 mm | 11 fps |
| 1.2 | Drilling the knee bone (Man age – 39) | Image registration 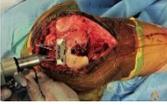 | 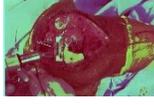 | 0.77mm | 8 fps | 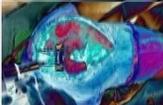 | 0.62 mm | 12 fps |
| | | Image overlay 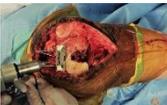 | 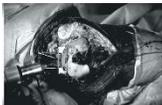 | 0.56 mm | 9 fps | 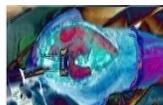 | 0.44 mm | 10 fps |
| | | If patient moves? 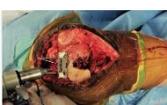 | 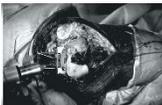 | 0.60 mm | 10 fps | 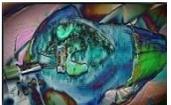 | 0.37 mm | 11 fps |
| | | If surgical instrument moves? 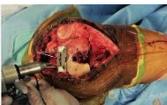 | 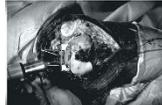 | 0.56 mm | 11 fps | 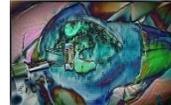 | 0.42 mm | 10 fps |
| 1.3 | Position of Implant (Man age – 40) | Image registration 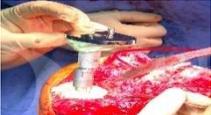 | 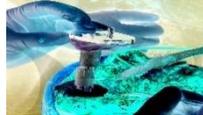 | 0.73 mm | 10 fps | 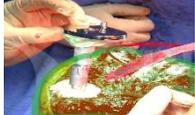 | 0.65 mm | 9 fps |
| | | Image overlay | | | | | | |

|  |  |  | 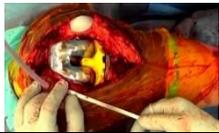 | 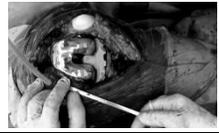 | 0.71 mm | 8 fps | 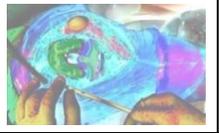 | 0.64 mm | 10 fps |
|---|---|---|---|---|---|---|---|---|---|
|  |  | If patient moves? | | | | | | | |
|  |  | | 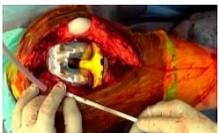 | 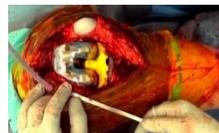 | 0.82 mm | 12 fps | 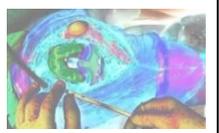 | 0.69 mm | 15 fps |
|  |  | If surgical instrument moves? | | | | | | | |
|  |  | | 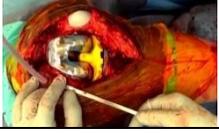 | 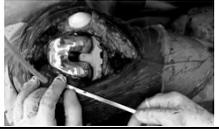 | 0.79 mm | 15 fps | 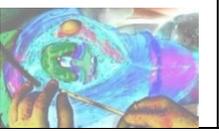 | 0.58 mm | 19 fps |
| 1.4 | Resurfacing the patella (Man age – 42) | Image registration | | | | | | | |
|  |  | | 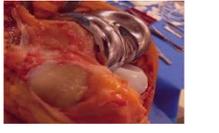 | 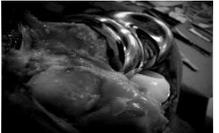 | 0.58 mm | 4 fps | 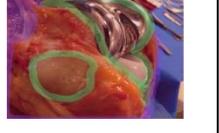 | 0.51 mm | 5 fps |
|  |  | Image overlay | | | | | | | |
|  |  | | 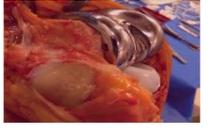 | 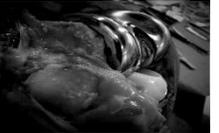 | 0.76 mm | 10 fps | 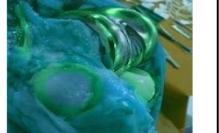 | 0.61 mm | 10 fps |
|  |  | If patient moves? | | | | | | | |
|  |  | | 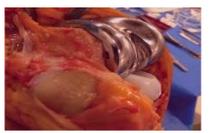 | 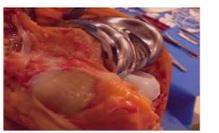 | 0.63 mm | 9 fps | 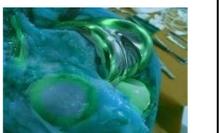 | 0.55 mm | 8 fps |
|  |  | If surgical instrument moves? | | | | | | | |
|  |  | | 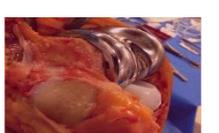 | 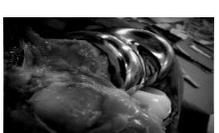 | 0.71 mm | 12 fps | 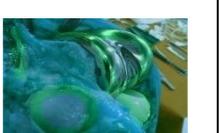 | 0.58 mm | 14 fps |
| 1.5 | Insertion of spacer to create smooth gliding surface (Man age – 41) | Image registration | | | | | | | |
|  |  | | 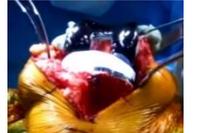 | 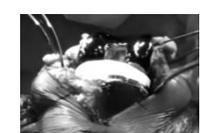 | 0.98 mm | 6 fps | 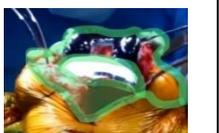 | 0.72 mm | 8 fps |
|  |  | Image overlay | | | | | | | |
|  |  | | 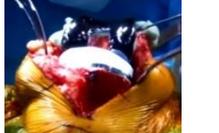 | 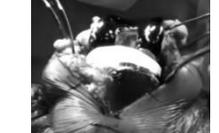 | 0.76 mm | 8 fps | 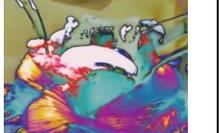 | 0.71 mm | 9 fps |

| | | If patient moves? | | | | | | |
|---|---|---|---|---|---|---|---|---|
| | | 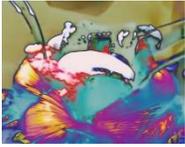 | 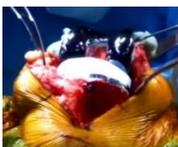 | 0.73 mm | 9 fps | 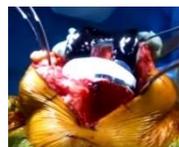 | 0.71 mm | 11 fps |
| | | If surgical instrument moves? | | | | | | |
| | | 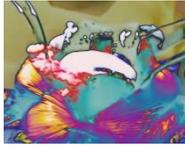 | 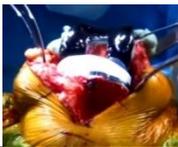 | 0.71 mm | 12 fps | 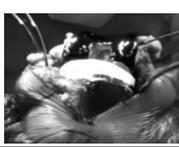 | 0.58 mm | 16 fps |

Table 2: Accuracy and Processing Time Results for Knee replacement (Sample 1: Adult Woman Age 35 to 45)

| Sample 1 Stages | Sample details Adult woman Age 35 to 45 | Original video | State of the Art | | | Proposed solution | | |
|---|---|---|---|---|---|---|---|---|
| | | | Processed sample | Accuracy by alignment error | Processing time (Frames per second) | Processed sample | Accuracy by overlay error | Processing time (Frames per second) |
| 1.1 | Opening Knee Bone (Woman age – 36) | Image registration | | | | | | |
| | | 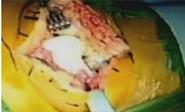 | 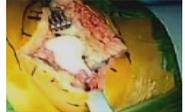 | 0.69 mm | 8 fps | 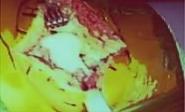 | 0.49 mm | 10 fps |
| | | Image overlay | | | | | | |
| | | 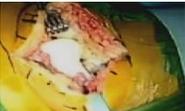 | 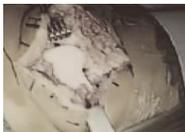 | 0.50 mm | 9 fps | 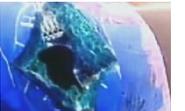 | 0.40 mm | 10 fps |
| | | If patient moves? | | | | | | |
| | | 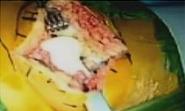 | 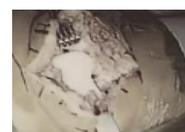 | 0.52 mm | 10 fps | 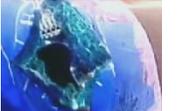 | 0.44 mm | 12 fps |
| | | If surgical instrument moves? | | | | | | |
| | | 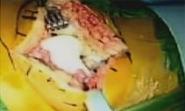 | 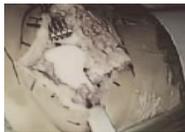 | 0.66 mm | 10 fps | 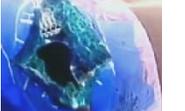 | 0.59 mm | 11 fps |
| 1.2 | Drilling the knee bone (Woman age – 39) | Image registration | | | | | | |
| | | 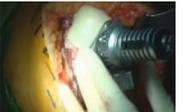 | 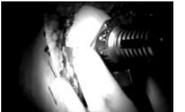 | 0.72 mm | 8 fps | 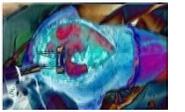 | 0.62 mm | 12 fps |
| | | Image overlay | | | | | | |

| | | | | | | | | | |
|---|---|---|---|---|---|---|---|---|---|
| | | | 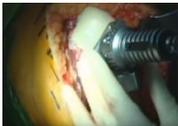 | 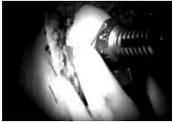 | 0.66 mm | 9 fps | 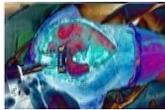 | 0.54 mm | 10 fps |
| | | If patient moves? | | | | | | | |
| | | | 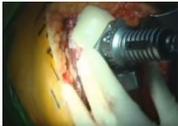 | 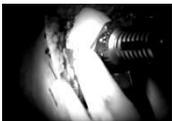 | 0.69 mm | 10 fps | 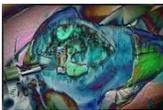 | 0.68 mm | 11 fps |
| | | If surgical instrument moves? | | | | | | | |
| | | | 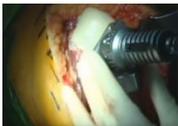 | 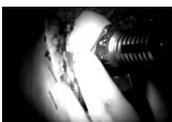 | 0.57 mm | 11 fps | 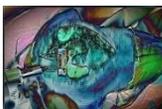 | 0.43 mm | 10 fps |
| 1.3 | Position of Implant (Woman age – 40) | Image registration | | | | | | | |
| | | | 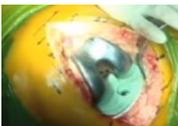 | 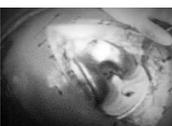 | 0.60 mm | 10 fps | 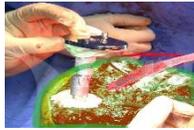 | 0.55 mm | 9 fps |
| | | Image overlay | | | | | | | |
| | | | 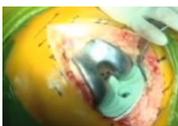 | 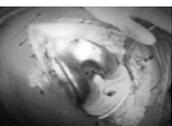 | 0.72 mm | 8 fps | 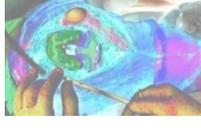 | 0.67 mm | 9 fps |
| | | If patient moves? | | | | | | | |
| | | | 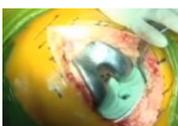 | 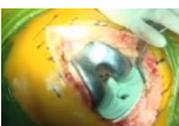 | 0.72 mm | 12 fps | 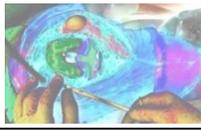 | 0.65 mm | 12 fps |
| | | If surgical instrument moves? | | | | | | | |
| | | | 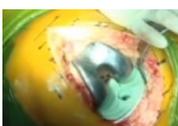 | 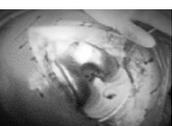 | 0.80 mm | 16 fps | 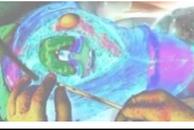 | 0.68 mm | 19 fps |
| 1.4 | Resurfacing the patella (Woman age – 41) | Image registration | | | | | | | |
| | | | 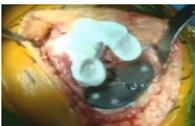 | 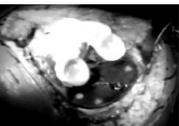 | 0.58 mm | 4 fps | 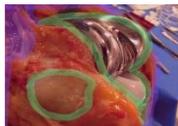 | 0.51 mm | 4 fps |
| | | Image overlay | | | | | | | |
| | | | 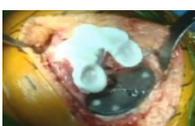 | 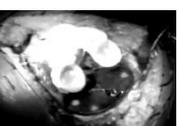 | 0.72 mm | 13 fps | 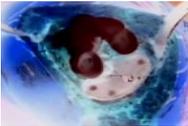 | 0.66 mm | 12 fps |
| | | If patient moves? | | | | | | | |

| | | | | 0.63 mm | 9 fps | 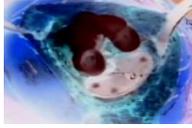 | 0.61 mm | 10 fps |
|---|---|---|---|---|---|---|---|---|
| | | 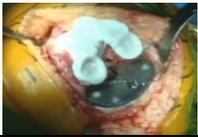 | 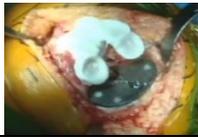 | | | | | |
| | | If surgical instrument moves? | | | | | | |
| | | 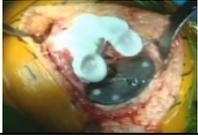 | 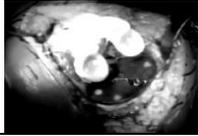 | 0.74 mm | 12 fps | 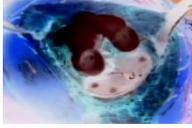 | 0.68 mm | 15 fps |
| 1.5 | Insertion of spacer to create smooth gliding surface (Woman age – 35) | Image registration | | | | | | |
| | | 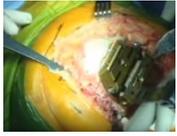 | 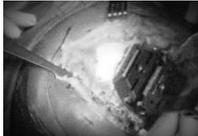 | 0.94 mm | 6 fps | 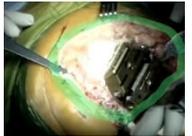 | 0.73 mm | 8 fps |
| | | Image overlay | | | | | | |
| | | 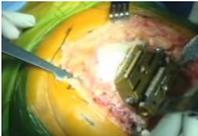 | 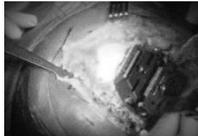 | 0.74 mm | 8 fps | 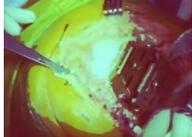 | 0.71 mm | 9 fps |
| | | If patient moves? | | | | | | |
| | | 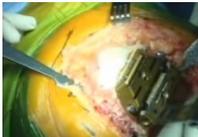 | 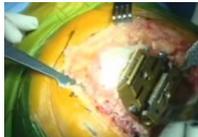 | 0.76 mm | 8 fps | 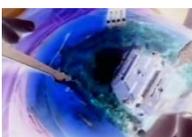 | 0.74 mm | 11 fps |
| | | If surgical instrument moves? | | | | | | |
| | | 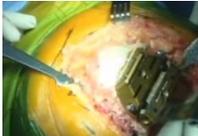 | 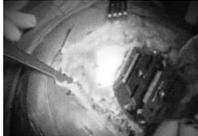 | 0.73 mm | 11 fps | 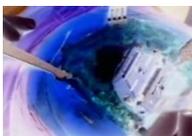 | 0.68 mm | 15 fps |

**Table 3: Accuracy and Processing Time Results for Knee replacement (Sample 1: Adult Age of persons both man and woman of age 20 to 30)**

| Sample 1 Stages | Sample details Adult man Age 20 to 30 | Original video | State of the Art | | | Proposed solution | | |
|---|---|---|---|---|---|---|---|---|
| | | | Processed sample | Accuracy by alignment error | Processing time (Frames per second) | Processed sample | Accuracy by overlay error | Processing time (Frames per second) |
| 1.1 | Opening Knee Bone (Man age – 22) | Image registration 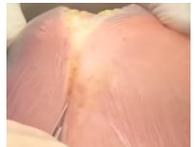 | 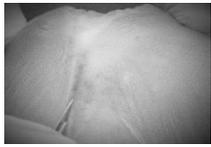 | 0.64 mm | 7 fps | 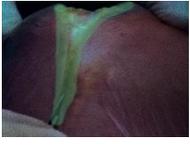 | 0.53 mm | 12 fps |
| | | Image overlay 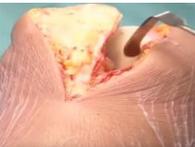 | 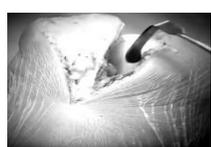 | 0.44 mm | 10 fps | 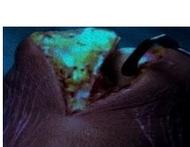 | 0.42 mm | 11 fps |
| | | If patient moves? 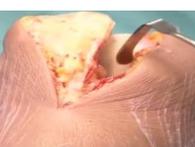 | 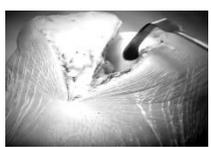 | 0.55 mm | 9 fps | 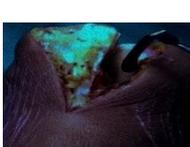 | 0.45 mm | 13 fps |
| | | If surgical instrument moves? 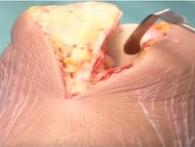 | 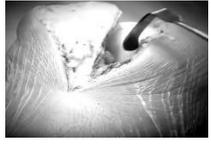 | 0.67 mm | 11 fps | 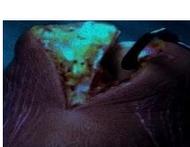 | 0.56 mm | 14 fps |
| 1.2 | Drilling the knee bone (Woman age – 23) | Image registration 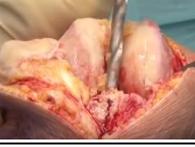 | 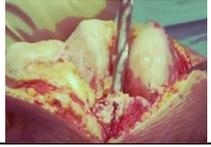 | 0.73 mm | 9 fps | 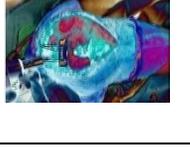 | 0.65 mm | 10 fps |
| | | Image overlay 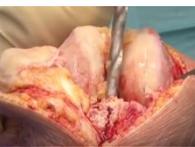 | 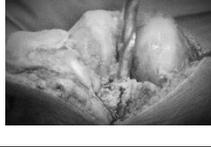 | 0.54 mm | 9 fps | 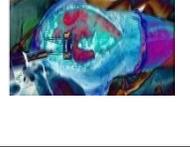 | 0.47 mm | 12 fps |
| | | If patient moves? 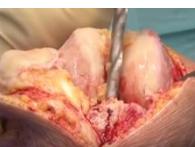 | 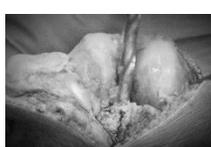 | 0.63 mm | 12 fps | 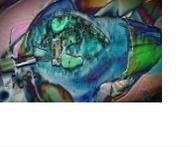 | 0.47 mm | 12 fps |

| | | If surgical instrument moves? | | | | | | |
|---|---|---|---|---|---|---|---|---|
| | | 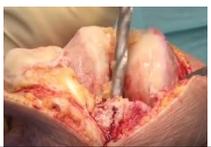 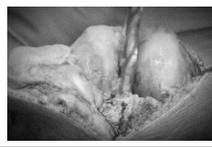 | 0.53 mm | 11 fps | 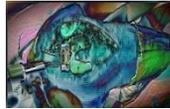 | 0.44 mm | 12 fps |
| 1.3 | Position of Implant (Man age – 29) | Image registration | | | | | |
| | | 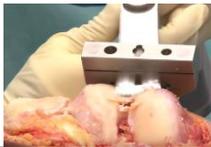 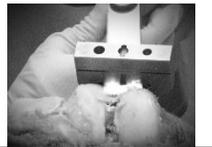 | 0.75 mm | 10 fps | 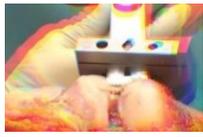 | 0.65 mm | 9 fps |
| | | Image overlay | | | | | |
| | | 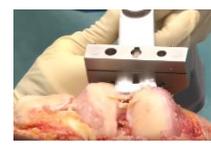 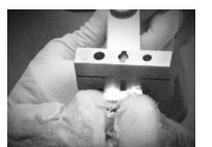 | 0.72 mm | 9 fps | 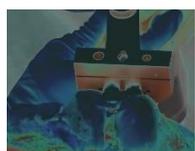 | 0.63 mm | 11 fps |
| | | If patient moves? | | | | | |
| | | 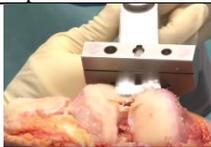 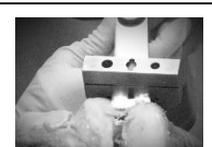 | 0.84 mm | 11 fps | 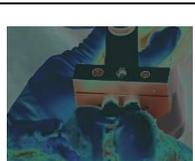 | 0.79 mm | 15 fps |
| | | If surgical instrument moves? | | | | | |
| | | 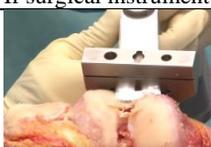 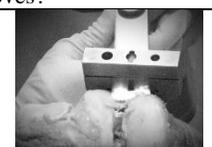 | 0.79 mm | 12 fps | 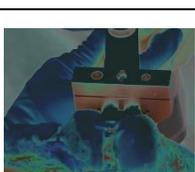 | 0.68 mm | 14 fps |
| 1.4 | Resurfacing the patella (Woman age – 26) | Image registration | | | | | |
| | | 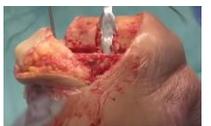 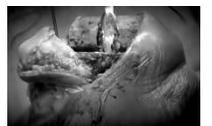 | 0.68 mm | 8 fps | 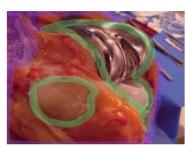 | 0.61 mm | 10 fps |
| | | Image overlay | | | | | |
| | | 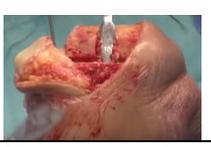 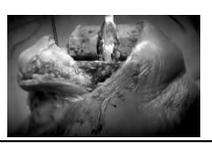 | 0.76 mm | 10 fps | 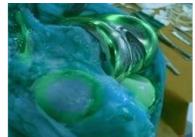 | 0.61 mm | 10 fps |
| | | If patient moves? | | | | | |
| | | 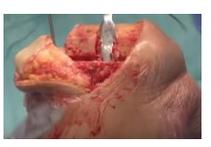 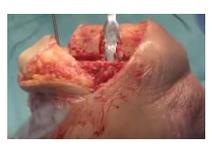 | 0.64 mm | 7 fps | 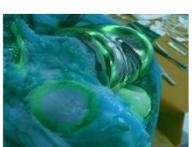 | 0.61 mm | 8 fps |
| | | If surgical instrument moves? | | | | | |
| | | 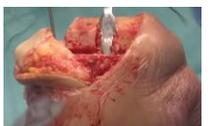 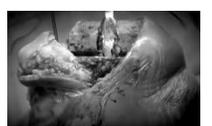 | 0.73 mm | 11 fps | 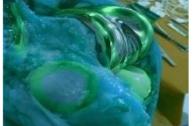 | 0.68 mm | 14 fps |

| 1.5 | Insertion of spacer to create smooth gliding surface (Man age – 30) | Image registration | | | | | | |
|---|---|---|---|---|---|---|---|---|
| | | 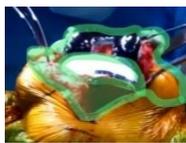 | 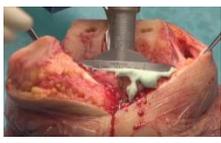 | 0.93 mm | 7 fps | 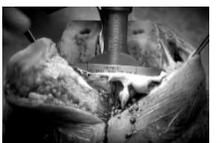 | 0.82 mm | 8 fps |
| | | Image overlay | | | | | | |
| | | 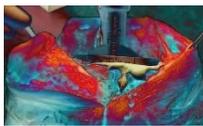 | 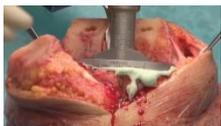 | 0.76 mm | 8 fps | 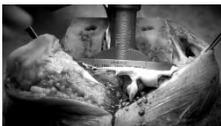 | 0.71 mm | 12 fps |
| | | If the patient moves? | | | | | | |
| | | 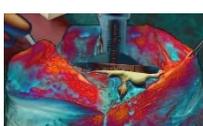 | 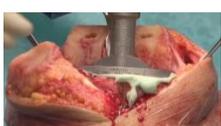 | 0.71 mm | 9 fps | 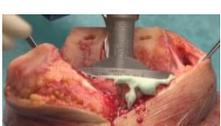 | 0.65 mm | 11 fps |
| | | If surgical instrument moves? | | | | | | |
| | | 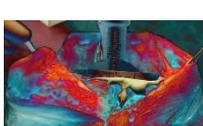 | 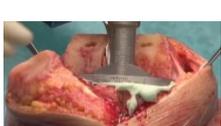 | 0.76 mm | 11 fps | 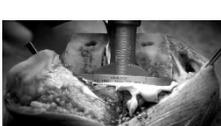 | 0.53 mm | 15 fps |

The green border in image registration of all samples indicates the success registering of images of both preoperative environment and intra-operative environments. The visualization of the blood vessels, nerve channels, and narrow surgical areas was enhanced with the use of the proposed algorithm. The improvement of the state-of-the-art solution and the proposed solution is presented in table 1,2 and 3.

The proposed system used the bidirectional maximum correntropy algorithm which helps to enhance the ICP algorithm and improves the misalignment problem faced by patients. The proposed system is markerless image registration techniques and solves the issues related to bulky trackers. Stereo based tracking with occlusion handling techniques helps to solve the occlusion problems caused by blood and surgical instruments. Furthermore, this proposed solution removes the non-Gaussian noise, outlier, impulse noise. Taking everything into account, the mix of the above technology and methods enormously improved the AR arrangement of visualization for knee medical procedures.

The accuracy of the proposed framework is obtained by figuring the distinction between the targeted overlay pixel in the picture and the genuine overlay pixel in the picture. Real images are from the bounding box area of a patient and targeted images are pre-processed data from the pre-operative environment. It has used a built-in function to calculate the disparity between left images and right images in terms of an image pixel. We successfully overcome the issues of state-of-the-art solution with 7.4 ~11.74 fps against 9.4 ~ 13.94 fps This research also improves alignment error 0.57 mm ~ 0.61 mm against 0.69 ~ 0.72 mm.

### Area of Improvement

The current research has proposed two solutions for occlusion handling and enhancing the ICP by ascertaining the optimal rigid transformation between two cloud points. This solution helps to eliminate alignment errors by removing the outliers and non-Gaussian noise. The main proposed solution is used in the matching of points in ICP. Among six variants of the Iterative closest point (ICP), the matching of the point is used for convergence and accelerating computation by finding the correspondence. ICP is very important using all points during the registration between two cloud meshes will slow down the convergence and even lead to the wrong pose. BiMCC is used in 3D-ICP registration for better alignment, and faster registration. This algorithm helps to eliminate the alignment process, maximize the overlapping parts and eliminate the registration outcomes trapped in local minima. BiMCC enhances the registration process, eliminates alignment error by ascertaining the optimal rigid transformation between two cloud points, and helps to improve the accuracy and processing time. BiMCC is reliable and favorable to eliminate alignment error by removing the outliers and non-Gaussian noise. Also, it is used to increase the speed of the alignment process by sampling the data and reducing the number of iterations of pose refinement. This algorithm is robust to the presence of noise, outliers, and partial differences. It helps to balance the penalty intensity automatically to adjust the trimming and matching ratio. Similarly, parameter estimation is solved to speed up computation and convergence by using two set points from the current correspondence of x and y cloud points. Another solution is added for occlusion handling using the Weight Least Square (WLS) algorithm. Which will first select the occluded point, define the neighborhood of the occluded point and interpolate by predicting the unknown values for those occluded points. This helps to fill the occluded point which is caused by instrument, blood, and other factors.

ICP is the most important algorithm in image registration and image alignment, which iteratively finds the point correspondence and updates the rigid transformation. One of the most important variants of ICP is the matching of the point where this variant is directly related to the robustness of the Iterative closest point algorithm. Different methods consider curvature, normal and other signals in point matching, however, these methods do not consider non-Gaussian noise, outliers, and partial difference. So, BiMCC is used in the proposed solution to enhance the registration process, eliminate alignment error by maximizing the overlapping parts, and eliminating the registration outcomes which are trapped in local minima. BiMCC is reliable and favorable which helps to ascertain the optimal rigid transformation between two cloud points. BiMCC is used to improve the speed of the alignment process by downsampling the data and reducing the number of iterations for posing refinement. This algorithm defines a bidirectional path for two cloud points and integrates the fixed-point optimization technique into an Iterative closest point framework where correspondence and transformation parameters are solved jointly. The main goal of this algorithm is to create correspondence and calculate the rigid transformation to ascertain the best alignment with two cloud points.

---

**Algorithm 2:** The ICP algorithm using BiMCC (**Bidirectional Maximum Correntropy**)

**Input**: Model Point cloud $\{\vec{x_i}\}_{i=1}^{N_x}$ and $\{\vec{y_j}\}_{j=1}^{N_y}$ and two set points $A = \{\vec{a_i}\}_{i=1}^{N}$ and $B = \{\vec{b_i}\}_{i=1}^{N}$

**Output**: Transformation R and $\vec{t}$, correspondence $\{i, d(j)_{j=1}^{N_y}\}$ and $\{c(i)_{i=1}^{N_x}, j\}$

**Start**:
1. Initialization: Set $R_0 = I_3$ and $\vec{t_0} = 0$, and randomly initialized $\sigma_0$
2. For k=1, 2, …, K do
3. Set up the correspondence $c_k(i)$ and $d_k(j)$ by nearest neighbour searching;
4. Compute the kernel $K_\sigma(e_k)$ by $R_{k-1}$ and $\vec{t_{k-1}}$
5. Find the error between two matched points
   $e_{xi} = (R_{xi} + \vec{T}) - \overrightarrow{y_{c(i)}}$
   Without good initial position of x, the position of y will be wrong at beginning. So, we need to fixed rotation matrix and then other translation points.
6. Two set points are used to accelerate the computation time and convergence.
   These two points can be defined as: -
   $$\vec{a_i} = \begin{cases} \vec{x_i}, & 1 \leq i \leq N_x, \\ x_{d_k(i)}, & N_x + 1 \leq i \leq N. \end{cases}$$

   $$\vec{b_i} = \begin{cases} y_{c_k(i)}, & 1 \leq i \leq N_x, \\ \vec{y_i}, & N_x + 1 \leq i \leq N. \end{cases}$$

   Two sets of points from current correspondence are used to speed up the convergence and computation. We have derived a modified error between two matched points.
   $Me_{xi} = (R\vec{a_i} + \vec{T}) - \vec{b_i}$
   $Me_{yi} = (R\overrightarrow{b_{d(j)}} + \vec{T}) - \vec{b_j}$
7. Solve the rotation matrix $R^*$

---

$R_k = R * R_{k-1}$, $\vec{t}_k = R * \vec{t}_{k-1} + \vec{t}^*$

8. Solve the translation vector $\vec{t}_0^*$
9. Update the transformation parameter $R_k$ and $\vec{t}_k$;
10. Compute transformation factor, rotation matrix and translation vector by combining forward distance and backward distance of two set points

$$M(\hat{v}_\sigma) = \frac{1}{N}(\sum_{i=1}^{N_x} \exp\left(\frac{-||R\vec{a_i}+\vec{T})-\vec{b_i}||^2}{2\sigma^2}\right) + \sum_{j=1}^{N_y} \exp\left(\frac{-||R\vec{b_{d(j)}}+\vec{T})-\vec{b_j}||^2}{2\sigma^2}\right))$$

11. Compute the MSE $Ee_k$;
   $EE(K) = \sum_{i=1}^{N} \exp[Mexi + Meyi]^2 / 2\sigma^2$
12. Set the kernel width $\sigma_k$;
13. Until $||e_k - e_{k-1}|| < \epsilon$.

## 5. Conclusion and Future Work

The proposed system consists of (ICP) Algorithm with (BiMCC). This algorithm helped to improve registration, alignment, maximize the overlapping parts between two cloud points, and eliminates the registration outcomes which are trapped in local minima. Furthermore, it has also removed the non-Gaussian noise, impulse noise, outliers and provided high registration accuracy. Rather than the Euclidean distance estimation, correntropy is utilized to quantify the shape comparability in the matching stages of the ICP variants, which further improved the upgraded calculation in posture refinement for precise and powerful outcomes. Besides, to speed up the convergence and computation two set points were derived from current correspondence in the matching phase, and modified errors between two matched points were calculated. Finally, with modified forward and backward direction during image registration, the iteration between two cloud points was decreased and processing time improved. By utilizing two variants of ICP (error estimation of points and matching points), the proposed framework was experimenting with a multi-paradigm numerical processing environment, i.e. MATLAB and has appeared to diminish the processing time and increment video accuracy. Alongside this, the proposed solution has considered the depth information which distinguished arrangement and made it clear the territories that should be cut. With the proposed solution, the processing time was improved to 9.45 ~11.25 fps against 12-15 fps, and alignment error has been decreased to 0.56 mm ~ 0.67 mm against 0.6 ~ 0.9 mm. See table 6. Among six metrics of ICP, we have only refined two metrics of ICP to improve image registration and processing time. See table 4. The future work can be done with the first metric, i.e. Selections of points. To solve the alignment error. Similarly, during pose refinement, a stereo matching algorithm should be refined to increase tracking quality with admissible losses at processing time. Stereo-based tracking in rigid bodies has still problems with camera arrangements, occlusion problems, and tracking accuracy. So, in the future work has been done in both image registration and pose refinement stages to increase the video accuracy and improve processing time.

Table 4. Comparison between the state of the art and the proposed solution

|  | **Proposed solution** | **State-of-art Solution** |
|---|---|---|
| **Name of the solution** | Bidirectional Maximum Correntropy | Iterative closest algorithm for image registration |
| **Applied Area** | Knee replacement surgery | Oral and maxillofacial surgery |
| **Accuracy** | 0.57 mm ~ 0.61 mm | 0.69 ~ 0.72 mm |
| **Processing time** | 7.4 ~11.74 fps | 9.4~13.94 fps |
| **Proposed Equation** | $EE(K) = \sum_{i=1}^{N} \exp[Mexi + Meyi]^2 / 2\sigma^2$ | $E(K) = \sum_{i=1}^{N} [(R\vec{x_i} + \vec{T}) - \vec{y}_{c(i)}]^2$ |
| **Contribution 1** | To speed up the computation and convergence we used two set points from the current correspondence of x and y cloud points. These points help to solve parameter estimation. | Convergence between rotational and translation vector is not considered in the state of the art which increases the processing time. |
| **Contribution 2** | Consists of enhancing Iterative Closest Point (ICP) Algorithm with Bidirectional Maximum Correntropy (BiMCC) which helps to improve registration, alignment, maximize the overlapping parts between two cloud points, and eliminates the registration outcomes which are trapped in local minima | Do not consider non-Gaussian noise, outlier, impulse noise, and overlapping parts. |